\documentclass{article}

\usepackage[preprint]{neurips_2026}

\usepackage[utf8]{inputenc} 
\usepackage[T1]{fontenc}    
\usepackage{hyperref}       
\usepackage{url}            
\usepackage{booktabs}       
\usepackage{amsfonts}       
\usepackage{nicefrac}       
\usepackage{microtype}      
\usepackage{xcolor}         
\usepackage{graphicx}
\usepackage{subcaption}
 \usepackage{amsmath}
\title{Trust Guided Decision Transformer}

\author{%
  Chainesh Gautam$^{1}$ \quad
  Raghuram Bharadwaj$^{1}$ \quad
  Chandramouli Kamanchi$^{2}$ \\
  \textbf{Pankaj Dayama$^{2}$ \quad Sumantha Mukherjee$^{2}$ \quad
  Kameshwaran Sampath$^{2}$} \\[4pt]
  $^{1}$International Institute of Information Technology Bangalore \quad
  $^{2}$IBM Research Bangalore \\[2pt]
  \texttt{\{chainesh.gautam, raghuram.bharadwaj\}@iiitb.ac.in} \\
  \texttt{chandramouli.kamanchi@ibm.com, \{pankajdayama, sumanm03, kameshwaran.s\}@in.ibm.com}
}

\begin{document}

\maketitle

\begin{abstract}
Decision Transformer performance degrades on long rollouts because the conditioning context drifts out of the training distribution. We show that this drift is visible through the model's own next state prediction error, which rises during rollout and stays elevated, giving a direct signal of when context has become unreliable. We introduce Trust Guided Decision Transformer (TGDT), which selects context before applying value guidance. At each step, TGDT evaluates several recent context suffixes using rolling next state prediction error, calibrated against held out offline data via split conformal prediction. It keeps only suffixes whose error stays within the calibrated threshold, then uses a frozen critic to choose the highest value action among the trusted suffixes. This reverses the order used by value only elastic selection, where the critic may choose an action generated from a context the model itself has flagged as unreliable. Experiments on D4RL navigation and locomotion tasks show that state prediction, critic guidance, and hard context reset each solve only part of the problem. TGDT reduces persistent high error runs and improves return over vanilla Decision Transformer, reset based context control, and value only context selection.
\end{abstract}

\section{Introduction}

Decision Transformer (DT)~\citep{chen2021decision} policies can degrade over long evaluation rollouts even when training has converged on offline data. We argue the cause is not the policy itself but the input the policy is being conditioned on. A DT is trained on context windows drawn from offline trajectories, but at evaluation time the context is generated by the model one step at a time. After enough self-generated steps, this rollout context can drift away from the offline distribution and the policy is asked to act on histories no training example resembles. The action it produces is then poorly constrained by the training objective. We call this failure mode \emph{rollout context mismatch}.

This mismatch has a measurable precursor. On an example D4RL Maze2D-medium (Figure 1a) task, the rolling next-state prediction error of a frozen probe rises above its held-out offline range during DT rollouts and stays elevated for long stretches, with the violation fraction increasing over episode time. The failure is not driven by isolated prediction spikes. It comes from persistent stretches in which the rollout context has left the regime where DT's training objective is informative. Because the probe is external to DT, this signal is a property of the rollout, not an artifact of the architecture we introduce later.

Most existing fixes for DT improve action quality given a context. Value-guided variants add a critic~\citep{yamagata2023qlearning,wang2024critic,zheng2025value}. Behavior-regularized variants keep actions near the offline distribution~\citep{fujimoto2021minimalist,tarasov2023revisiting}. Adaptive-context methods change the amount of trajectory history used at test time to improve stitching~\citep{wu2023elastic}. These methods largely ask which action is best given a context. Our observation suggests that a different question must come first: which contexts can be trusted to produce actions worth ranking at all? Critic value does not tell us whether the context that produced an action is reliable. Ranking actions generated from an unreliable context can mean choosing among options the model itself has already flagged as suspect.

We introduce Trust Guided Decision Transformer (TGDT), which separates context reliability from action value at the decision level. At each step, TGDT queries the model on several context-suffix lengths, maintains a rolling next-state prediction error per suffix, rejects suffixes whose recent error has exceeded a held-out offline threshold, and only then asks the critic to rank actions among the surviving suffixes.

TGDT shares the premise of adaptive-context methods that context length should be chosen online, but it uses a different selection criterion. In a controlled ablation using the same trained model, frozen critic, and candidate suffix set, critic-only suffix selection ranks every candidate by critic value and executes the winner. TGDT removes candidates whose recent prediction error exceeds the reliability threshold, then ranks only among the rest. On Maze2D-medium, this reordering reduces the longest above-threshold run by approximately \(8\times\) and improves normalized return. The contribution is the order of operations, not a new critic or a larger model.

TGDT uses a learned next-state predictor, but it is not model-based offline reinforcement learning. The predictor is never used to roll out trajectories, generate synthetic transitions, or optimize a policy through model rollouts~\citep{yu2020mopo,kidambi2020morel,yu2021combo}. It is used only as a per-step reliability sensor for the realized rollout context. The policy remains a return-conditioned sequence model, and the frozen critic is used only to rank candidate actions produced by trusted context suffixes.

Our contributions are as follows:
\begin{enumerate}
    \item We identify rollout context mismatch as a DT failure mode with a measurable precursor. Persistent rolling next-state prediction error precedes return degradation, and the signal is observable from a frozen external probe, which establishes it as a property of the rollout rather than an artifact of a specific architectural choice.

    \item We show that training-side improvements alone are insufficient for reliability. Auxiliary next-state prediction, frozen IQL critic guidance, and a bounded behavior-regularized residual action head each improve return, but the resulting policies still produce long above-threshold runs of high prediction error during evaluation.

    \item We introduce TGDT, a trust-filtered context-selection rule that conditions the critic's ranking on a held-out reliability threshold derived from offline data. On D4RL Maze2D, AntMaze, and MuJoCo locomotion, TGDT improves over the same trained model evaluated with full context, hard reset, and critic-only suffix selection.
\end{enumerate}

\section{Related work}
\label{sec:related}

Our method builds on three lines of work. The first studies Decision Transformers and value guided sequence models for offline reinforcement learning. The second studies behavior regularization and offline support constraints. The third studies adaptive context length and data derived reliability thresholds. TGDT differs from these lines by asking a different question. Before ranking actions by value, the agent must decide whether the context that produced those actions is reliable.

\paragraph{Decision Transformers and value guided action generation.}
Decision Transformer formulates offline reinforcement learning as return conditioned sequence modeling over states, actions, and returns to go~\citep{chen2021decision}. Trajectory Transformer similarly models offline trajectories as sequences and uses the learned model for planning~\citep{janner2021offline}. These methods avoid explicit policy improvement during training, but pure sequence imitation can struggle when the requested return is poorly supported by the dataset or when the policy must stitch useful parts of different trajectories. Several recent methods add value information to Decision Transformers. QDT uses Q learning to improve return conditioning and compensate for limitations of pure supervised sequence modelling~\citep{yamagata2023qlearning}. CGDT uses a critic to align target returns with expected returns~\citep{wang2024critic}. ACT introduces advantage conditioning to bring dynamic-programming-style improvement into DT training~\citep{gao2024act}. VDT uses value functions to guide Decision Transformer training through advantage weighting and behavior regularization~\citep{zheng2025value}. \citet{brandfonbrener2022when} analyse when return-conditioned supervised learning succeeds and identify regimes in which it can be unreliable, which complements our diagnosis that the conditioning context itself can become unreliable during evaluation rollouts.

\paragraph{Behavior regularization in offline reinforcement learning.}
TGDT uses behavior regularization to keep candidate actions close to the offline data. This follows a standard principle in offline reinforcement learning. BCQ restricts policy improvement to actions supported by the behavior distribution~\citep{fujimoto2019off}. TD3+BC adds a behavior cloning term to the actor objective~\citep{fujimoto2021minimalist}. ReBRAC revisits this recipe and shows that careful behavior regularization remains a strong offline reinforcement learning baseline~\citep{tarasov2023revisiting}. IQL avoids explicit maximization over unseen actions during critic training~\citep{kostrikov2021offlinereinforcementlearningimplicit}. AWR learns a policy through advantage weighted regression~\citep{peng2019advantage}. We do not claim novelty in these ingredients. Their role is to keep TGDT's generated actions near the offline action distribution, so that next state prediction error remains interpretable as a signal of context mismatch rather than arbitrary action drift.

\paragraph{Adaptive context length at inference.}
Elastic Decision Transformer (EDT) adapts the amount of history used by a Decision Transformer at test time to improve trajectory stitching~\citep{wu2023elastic}. TGDT shares the idea that context length should be chosen online rather than fixed in advance. The difference is the selection signal. EDT motivates value-based history selection, while TGDT uses recent next-state prediction error to decide which context suffixes are reliable before applying critic ranking. Thus, TGDT treats context selection as a reliability problem before it treats it as a value-ranking problem.

\paragraph{Conformal prediction and held out reliability thresholds.}
The threshold in TGDT is computed from held out prediction errors. This uses the same order statistic as split conformal prediction~\citep{vovk2005algorithmic,angelopoulos2023conformal}. In the exchangeable setting, split conformal prediction gives finite sample coverage. Closed loop reinforcement learning rollouts are not exchangeable with teacher forced validation trajectories because future states depend on the policy's previous actions. Recent work studies conformal prediction under covariate shift, adaptive thresholds, and non exchangeable data~\citep{tibshirani2019conformal,gibbs2021adaptive,barber2023conformal}. TGDT does not claim an online coverage guarantee. It uses the held out quantile as a reliability threshold for context selection and evaluates that threshold by its effect on persistent prediction error and return.

\paragraph{Dynamics models in offline reinforcement learning.}
TGDT uses a next state prediction head, but it is not a model based offline reinforcement learning method. Model based offline reinforcement learning methods such as MOPO, MOReL, and COMBO train dynamics models and use them for rollout generation, uncertainty penalties, or conservative policy optimization~\citep{yu2020mopo,kidambi2020morel,yu2021combo}. TGDT does not roll out the learned dynamics model and does not optimize a policy inside the model. The prediction head is used only as a reliability sensor for the context. The policy remains a return conditioned sequence model, and the critic ranks only actions proposed by trusted context suffixes.
\section{Method}
\label{sec:method}

The introduction defined rollout context mismatch as the failure mode in which a Decision Transformer conditions on histories its training distribution does not contain, and identified persistent next state prediction error as a measurable precursor. This section turns that observation into a method.

We call the method Trust Guided Decision Transformer (TGDT). Section~\ref{sec:method-diagnosis} establishes that rollout context mismatch can be detected online from the next state prediction error of an external probe, even for a vanilla DT that has no internal prediction head. Section~\ref{sec:method-training} adds three components to the DT itself, namely a frozen IQL critic, a behavior regularized residual action head, and an internal reliability head, so that the runtime rule does not need an external probe. Sections~\ref{sec:method-threshold} and~\ref{sec:method-evaluation} derive the reliability threshold and the trust filtered selection rule. Throughout, the central design principle is that value should rank actions only after context reliability has been checked.

\subsection{Detecting rollout context mismatch with a transition probe}
\label{sec:method-diagnosis}

A vanilla DT~\citep{chen2021decision} maps a length \(K\) context \(h_t^{(K)}=(R_{t-K:t},s_{t-K:t},a_{t-K:t-1})\) to an action \(\hat a_t = \pi_\theta(h_t^{(K)},R_t,s_t)\). The model has no internal estimate of how reliable its current context is. To probe the rollout for context mismatch without modifying the trained model, we use an external transition predictor.

The probe is a small network \(g_\omega\) that maps a DT context and an executed action to a next state prediction \(\hat s_{t+1}^{\mathrm{probe}} = g_\omega(h_t^{(K)},R_t,s_t,\hat a_t)\). It is trained on the offline dataset to minimize \(\|s_{t+1}-g_\omega(\cdot)\|^2\) under teacher forcing, and frozen before evaluation. After the environment reveals \(s_{t+1}\), the rolling probe error over a window \(K_e\) is
\begin{equation}
S_t^{\mathrm{probe}} = \frac{1}{K_e d_s}\sum_{j=t-K_e+1}^{t}\|\hat s_{j+1}^{\mathrm{probe}}-s_{j+1}\|^2.
\label{eq:probe-score}
\end{equation}
The probe is used only for diagnosis. TGDT's runtime rule uses the model's own internal reliability head, introduced in Section~\ref{sec:method-training}. Appendix~\ref{app:probe-internal-agreement} shows that the two signals identify similar persistent high error regimes on Maze2D, with Pearson correlation \(r \in [0.82,0.88]\) and threshold agreement at least \(0.86\). This supports using the internal head as the runtime reliability signal once the DT is trained.

Section~\ref{sec:exp-precursor} tests whether vanilla DT rollouts enter regions where \(S_t^{\mathrm{probe}}\) rises above its typical offline range and stays there for long stretches. The question is not whether prediction error spikes during failure, which would be unsurprising, but whether persistent high error \emph{precedes} return collapse rather than coinciding with it. Two observations follow if it does. First, the reliability of the current rollout context is observable through prediction error, even when vanilla DT does not measure it internally. Second, an intervention that reads this signal online can act before the failure compounds. The remainder of this section turns these observations into a method.

\subsection{Training a value guided DT with a reliability head}
\label{sec:method-training}

The probe in Section~\ref{sec:method-diagnosis} reveals the problem but is not part of the policy. The runtime rule needs the DT itself to produce a next state prediction for every candidate suffix. It also needs action proposals that remain close to the data, so that the reliability signal reflects context drift rather than action drift, and a way to rank candidate actions. We obtain these from three model components, all trained jointly.

\paragraph{Frozen offline critic.}
A critic \(Q_\phi(s,a)\) is trained on the offline dataset with implicit Q learning~\citep{kostrikov2021offlinereinforcementlearningimplicit}. The critic remains frozen for the rest of training and at evaluation. Implicit Q learning is chosen because it avoids explicit maximization over unseen actions during critic training. The critic plays two roles. During DT training it provides a small value signal that biases the action head toward higher value behavior. During evaluation it ranks DT generated candidate actions across a small finite set of context suffixes. It is never used to search the continuous action space.

\paragraph{Behavior regularized action head.}
Without constraints on the action distribution, the reliability signal in Section~\ref{sec:method-threshold} loses its meaning. If the action head drifts far from the dataset, rolling prediction error reflects drift in actions rather than drift in context, and the trust filter would reject reliable contexts that happen to be paired with out of distribution actions. We therefore predict the action as a base prediction plus a bounded residual,
\begin{equation}
a_t^{\mathrm{base}} = \pi_\theta^{\mathrm{base}}(z_t),\quad
\Delta_t = \delta_{\max}\tanh(f_\psi(z_t)),\quad
\hat a_t = \tanh(a_t^{\mathrm{base}}+\Delta_t),
\end{equation}
where \(z_t\) is the action token's hidden state and \(\delta_{\max}\) is a small constant. The action loss is
\begin{equation}
\mathcal L_{\mathrm{act}} = w_t\|\hat a_t-a_t\|^2 - \lambda_Q Q_\phi(s_t,\hat a_t) + \beta\|\Delta_t\|^2,
\label{eq:loss-act}
\end{equation}
following the behavior regularized improvement template of TD3+BC~\citep{fujimoto2021minimalist} and ReBRAC~\citep{tarasov2023revisiting}. The first term keeps \(\hat a_t\) near the dataset action \(a_t\), optionally weighted by a normalized advantage \(w_t\) in the style of advantage weighted regression~\citep{peng2019advantage}. The second term applies the small critic improvement. The third term penalizes large residuals beyond what \(\delta_{\max}\) already enforces. With small \(\lambda_Q\) and \(\delta_{\max}\), the BC term and the residual bound dominate, and the critic chooses better actions within a near data neighborhood rather than searching freely over actions. Although this correction is local at each step, its effect can compound over long rollouts.

\paragraph{Internal reliability head.}
The DT also predicts the next state directly from the action token's representation, \(\hat s_{t+1} = g_\theta(z_t)\). The state prediction loss is the normalized mean squared error
\begin{equation}
\mathcal L_{\mathrm{state}} = \frac{1}{d_s}\|\hat s_{t+1}-s_{t+1}\|^2.
\label{eq:loss-state}
\end{equation}
At inference, this head produces \(\hat s_{t+1}\) before \(s_{t+1}\) is observed, which is the property the runtime rule requires. The full training objective is
\begin{equation}
\mathcal L = w_t\|\hat a_t-a_t\|^2 - \lambda_Q Q_\phi(s_t,\hat a_t) + \beta\|\Delta_t\|^2 + \lambda_s \cdot \mathcal L_{\mathrm{state}}.
\label{eq:loss-total}
\end{equation}
After training, the DT model emits both an action and a next state prediction from the same context. The probe of Section~\ref{sec:method-diagnosis} is no longer needed.

\subsection{Held out reliability threshold}
\label{sec:method-threshold}

The next state prediction error has no universal scale. A value of \(0.05\) may be unremarkable on one dataset and clearly anomalous on another, depending on environment stochasticity and how well the predictor can fit the offline data. The threshold above which an error is considered anomalous must therefore be derived from the offline data itself.

We use a held out split of the offline trajectories. On this split, the trained model is run under teacher forcing and produces a per step prediction error \(e_t = \tfrac{1}{d_s}\|\hat s_{t+1}-s_{t+1}\|^2\). Individual errors are noisy, both because environment dynamics are stochastic and because the predictor is imperfect even on offline data. A persistent rise is the signal we care about, not a single spike. We therefore work with the rolling mean over a window of length \(K_e\),
\begin{equation}
S_t = \frac{1}{K_e}\sum_{j=t-K_e+1}^{t}e_j,
\label{eq:rolling-score}
\end{equation}
which suppresses isolated spikes and emphasizes persistent error. Let \(\{S_i^{\mathrm{val}}\}_{i=1}^{n}\) be the rolling scores collected across the held out split. For a risk level \(\alpha\), we set
\begin{equation}
\tau_\alpha = S^{\mathrm{val}}_{(\lceil (n+1)(1-\alpha)\rceil)},
\label{eq:tau}
\end{equation}
the \(\lceil (n+1)(1-\alpha)\rceil\)-th order statistic. We use \(\alpha=0.05\) in all experiments, so \(\tau_\alpha\) is the \(95\)th percentile of rolling errors observed under offline contexts.

Equation~\ref{eq:tau} uses the same order statistic as split conformal calibration~\citep{vovk2005algorithmic, 10.5555/1886351.1886356}. In the exchangeable setting, this quantile gives a finite sample tail guarantee. Our closed loop setting is not exchangeable, so we do not use the threshold as a formal coverage guarantee. Three sources of non exchangeability apply. Test scores are temporally correlated through the rolling window. The rollout state distribution depends on the policy's own past actions. The action distribution at evaluation differs from the dataset's behavior policy that produced the validation scores. We therefore use \(\tau_\alpha\) as a \emph{held out reliability threshold}, an empirical scale grounded in offline data, and judge it by its operational consequences in Section~\ref{sec:exp-execution}. Restoring approximate coverage with non exchangeable conformal machinery~\citep{barber2023conformal, gibbs2021adaptive} is a natural extension we leave to future work.

\subsection{Trust filtered context selection}
\label{sec:method-evaluation}

The runtime rule combines the trained components into a single decision per step. The DT proposes candidate actions from several context suffix lengths. Suffixes whose recent rolling score exceeds the threshold are rejected. The critic ranks actions only among the surviving suffixes. We now state the rule precisely.

Let \(\mathcal L\subseteq\{1,\ldots,K\}\) be a small set of candidate context lengths. 
For each \(L\in\mathcal L\), the suffix context \(h_t^{(L)}\) contains the last \(L\) steps of the realized history. The model is queried independently on each suffix to produce a candidate action,
\begin{equation}
\hat a_t^{(L)} = \pi_\theta(h_t^{(L)},R_t,s_t).
\label{eq:candidate}
\end{equation}
We use L = {1, 5, 10, K} as the default. Appendix \ref{app:context-length-usage} reports the per-suffix selection frequencies and shows that TGDT uses all four lengths in practice rather than collapsing to the shortest one. The cost per step is \(|\mathcal L|\) extra forward passes \ref{app:runtime-overhead}, independent of the rollout length. These candidate actions are used only for value ranking inside the rule below. 

Each suffix maintains its own rolling queue of prediction errors. At decision time \(t\), the queue for suffix \(L\) holds the most recent \(K_e\) errors observed under that suffix, all from steps strictly before \(t\). The corresponding score is
\begin{equation}
S_{t-1}^{(L)} = \frac{1}{K_e}\sum_{j=t-K_e}^{t-1}e_j^{(L)}.
\label{eq:per-suffix-score}
\end{equation}
The per suffix errors are not counterfactual rollouts. They are different views of the same realized trajectory under different context truncations. The trusted set at step \(t\) is
\begin{equation}
\mathcal A_t = \{L\in\mathcal L \,:\, S_{t-1}^{(L)}\le \tau_\alpha\}.
\label{eq:trusted-set}
\end{equation}
Among trusted suffixes, the rule selects
\begin{equation}
\boxed{\;L_t^* = \arg\max_{L\in\mathcal A_t} Q_\phi\!\left(s_t,\hat a_t^{(L)}\right),\;}
\label{eq:selection}
\end{equation}
and executes \(a_t = \hat a_t^{(L_t^*)}\). If \(\mathcal A_t\) is empty, the rule falls back to \(L=1\), the shortest context. The fallback rarely fires once a calibrated threshold is in place, but it ensures the rule is always defined.

After the action is executed and the environment reveals \(s_{t+1}\), the reliability head predicts the next state from each suffix context under the executed action,
\begin{equation}
\tilde s_{t+1}^{(L)} = g_\theta(h_t^{(L)},R_t,s_t,a_t).
\end{equation}
The realized suffix error is
\begin{equation}
e_t^{(L)} = \frac{1}{d_s}\|\tilde s_{t+1}^{(L)}-s_{t+1}\|^2.
\end{equation}
This compares all suffixes on the same observed transition rather than on counterfactual actions that were not executed. These errors update the scores \(S_t^{(L)}\), which are used at the next decision. The score that gates suffix \(L\) at step \(t\) uses only errors from steps before \(t\), and the realized \(s_{t+1}\) is used only after \(a_t\) has been committed. The rule is therefore causal and avoids future leakage.

\paragraph{Comparison to critic-only suffix selection.}
A critic-only variant drops the trust filter and selects
$
\arg\max_{L\in\mathcal L}
Q_\phi(s_t,\hat a_t^{(L)}).
$
This is our closest controlled baseline. It uses the same trained model, the same frozen critic, and the same candidate suffix set as TGDT, but removes the reliability filter. It also captures the value-ranking idea behind adaptive context-selection methods such as EDT~\citep{wu2023elastic}, without treating EDT itself as the matched ablation. Critic-only selection ranks all candidates by critic value and accepts the highest-valued suffix. TGDT first rejects candidates whose recent rolling error has exceeded \(\tau_\alpha\), and only then ranks the trusted suffixes. When the model's own prediction error has flagged a context as unreliable, TGDT refuses to consider actions produced from that context, regardless of how the critic scores them. The experiments in Section~\ref{sec:exp-execution} test whether this ordering explains the gap left by vanilla DT, hard reset, and critic-only suffix selection.

\section{Experiments}
\label{sec:experiments}

We evaluate TGDT on D4RL~\citep{fu2020d4rl} and ask three questions. First, does vanilla DT develop a measurable reliability failure during closed-loop rollout? Second, do training-side improvements remove this failure on their own? Third, does reliability filtering improve over critic-only suffix selection when both methods use the same trained model, the same frozen critic, and the same candidate suffix set?

\subsection{Setup}
\label{sec:exp-setup}

\paragraph{Datasets and Baselines.}
We evaluate on five D4RL \citep{fu2020d4rl} domains, namely Maze2D, AntMaze, MuJoCo locomotion, Adroit, and Kitchen, covering long-horizon goal reaching, sparse-reward navigation, dense-reward continuous control, dexterous manipulation, and multi-step task chains. We compare against representative value-based offline RL methods (BEAR~\citep{kumar2019stabilizingoffpolicyqlearningbootstrapping}, BCQ~\citep{fujimoto2019off}, CQL~\citep{kumar2020conservative}, IQL~\citep{kostrikov2021offlinereinforcementlearningimplicit}, MoRel~\citep{kidambi2020morel}, O-RL~\citep{brandfonbrener2021offlinerloffpolicyevaluation}, COMBO~\citep{yu2021combo}) and conditional sequence modeling methods (BC, DT~\citep{chen2021decision}, StAR~\citep{Shang_2022}, GDT~\citep{hu2023graphdecisiontransformer}, CGDT~\citep{wang2024critic}, DC~\citep{kim2024decisionconvformerlocalfiltering}, QDT~\citep{yamagata2023qlearning}, EDAC~\citep{an2021uncertainty}, D-QL~\citep{wang2022diffusion}, DD~\citep{ajay2022conditional},VDT~\citep{zheng2025value}). Baseline scores are sourced from the best results published in their respective papers, with missing entries reproduced from our own runs to ensure a fair comparison.

\paragraph{Evaluation protocol.}
For each task and method we train three independent seeds and evaluate 100 episodes per seed under each execution mode, with paired evaluation seeds across modes within a trained seed. Reported numbers for our methods are mean and standard error of normalized rewards across the three seeds. This protocol applies throughout the section unless otherwise stated. Full hyperparameters, advantage-weighting variants, and compute details are in Appendix~\ref{app:training-hparam-sensitivity}

\paragraph{Trained variants and execution modes.}
Four training variants isolate each ingredient of the full loss in Equation~\ref{eq:loss-total}, namely vanilla \textbf{DT}~\citep{chen2021decision}, \textbf{DT + SP} (state-prediction head only), \textbf{DT + Critic} (residual action head with critic regularization only), and \textbf{DT + Critic + SP} (full TGDT model). On the trained full model we evaluate four execution modes, namely \texttt{none} (full context, no selection), \texttt{hard} (reset to \(L=1\) on threshold violation), \texttt{critic-only} (rank suffixes by $Q_{\phi}$ without filter, capturing the value-only selection idea behind adaptive context-selection methods such as \citep{wu2023elastic}), and TGDT \ref{eq:selection}. Training variants lacking the components required for trust filtering are evaluated only under \texttt{none}. 

\subsection{Main D4RL results}
\label{sec:exp-main}

Table~\ref{tab:main-d4rl} reports normalized D4RL scores across the benchmark domains. The pattern matches the structure of our diagnosis. TGDT improves most on tasks where the rollout horizon allows context mismatch to accumulate before episode end. The gain is largest on Maze2D, where long-horizon stitching gives unreliable contexts many opportunities to corrupt later decisions. On AntMaze medium-diverse, TGDT roughly doubles the score of the best sequence-modeling baseline (VDT, 30.0 → 60.0), although \citep{kostrikov2021offlinereinforcementlearningimplicit} still leads among value-based methods (70.0). The pattern is consistent with a sparse-reward setting where many candidate sub-trajectories must be assembled to reach the goal.

On dense-reward MuJoCo locomotion, TGDT remains competitive with strong value-guided sequence-modeling baselines but the gain is smaller, since each transition contributes substantial reward and the value of preventing late-episode failures from context drift is reduced. The same explanation applies to Adroit and Kitchen, where TGDT improves over the prior best on most tasks but with smaller absolute margins than on Maze2D and AntMaze. TGDT is most useful when the rollout is long enough for context drift to accumulate and episode reward depends on decisions made after that drift would have begun. When neither condition is sharply met, TGDT does not hurt but the gain over the best value-guided baseline is modest.

\begin{table*}[t]
\centering
\scriptsize
\setlength{\tabcolsep}{4.2pt}
\renewcommand{\arraystretch}{1.08}
\caption{Offline D4RL performance. Scores are normalized D4RL scores. TGDT denotes our trust-filtered execution rule. See Section~\ref{sec:exp-setup} for the evaluation protocol and baseline sourcing.}
\label{tab:main-d4rl}
\resizebox{\textwidth}{!}{
\begin{tabular}{l||ccccc|cccccccc}
\toprule
\textbf{Dataset}
& \multicolumn{5}{c|}{\textbf{Value-Based Methods}}
& \multicolumn{8}{c}{\textbf{Conditional Sequence Modeling Methods}} \\
\midrule

\textbf{Gym Tasks}
& \textbf{BEAR} & \textbf{BCQ} & \textbf{CQL} & \textbf{IQL} & \textbf{MoRel}
& \textbf{BC} & \textbf{DT} & \textbf{StAR} & \textbf{GDT} & \textbf{CGDT} & \textbf{DC} & \textbf{VDT} & \textbf{TGDT} \\
\midrule
halfcheetah-medium-replay-v2
& 38.6 & 34.8 & 37.5 & 44.1 & 40.2
& 36.6 & 36.6 & 36.8 & 40.5 & 40.4 & 41.3 & 39.4$\pm$2.0 & \textbf{45.1}$\pm$1.7 \\
hopper-medium-replay-v2
& 33.7 & 31.1 & 95.0 & 92.1 & 93.6
& 18.1 & 82.7 & 29.2 & 85.3 & 93.4 & 94.2 & \textbf{96.0}$\pm$1.9 & 95.5$\pm$2.2 \\
walker2d-medium-replay-v2
& 19.2 & 13.7 & 77.2 & 73.7 & 49.8
& 32.3 & 79.4 & 39.8 & 77.5 & 78.1 & 76.6 & 82.3$\pm$2.1 & \textbf{83.9}$\pm$2.7 \\
\midrule
halfcheetah-medium-v2
& 41.7 & 41.5 & 44.0 & \textbf{47.4} & 42.1
& 42.6 & 42.6 & 42.9 & 42.9 & 43.0 & 43.0 & 43.9$\pm$0.7 & 45.9$\pm$0.9 \\
hopper-medium-v2
& 52.1 & 65.1 & 58.5 & 63.8 & 95.4
& 52.9 & 67.6 & 59.5 & 77.1 & 96.9 & 92.5 & 98.3$\pm$0.1 & \textbf{99.2}$\pm$0.3 \\
walker2d-medium-v2
& 59.1 & 52.0 & 72.5 & 79.9 & 77.8
& 75.3 & 74.0 & 73.8 & 76.5 & 79.1 & 79.2 & 81.6$\pm$1.7 & \textbf{82.7}$\pm$1.1 \\
\midrule
halfcheetah-medium-expert-v2
& 53.4 & 69.6 & 91.6 & 86.7 & 53.3
& 55.2 & 86.8 & 93.7 & 93.2 & 93.6 & 93.0 & 93.9$\pm$0.1 & \textbf{96.8}$\pm$0.1 \\
hopper-medium-expert-v2
& 96.3 & 109.1 & 105.4 & 91.5 & 108.7
& 52.5 & 107.6 & 111.1 & 111.1 & 107.6 & 110.4 & 111.5$\pm$3.8 & \textbf{112.9}$\pm$0.6 \\
walker2d-medium-expert-v2
& 40.1 & 67.3 & 108.8 & 109.6 & 95.6
& 107.5 & 108.1 & 109.0 & 107.7 & 109.3 & 109.6 & 110.4$\pm$0.9 & \textbf{115.1}$\pm$1.2 \\
\midrule
\textit{Average}
& 48.2 & 53.8 & 77.6 & 76.5 & 72.9
& 52.6 & 76.2 & 66.2 & 79.1 & 82.4 & 82.2 & 84.1 & \textbf{86.3} \\

\midrule\midrule
\textbf{Adroit Tasks}
& \textbf{BEAR} & \textbf{BCQ} & \textbf{CQL} & \textbf{IQL} & \textbf{MoRel}
& \textbf{EDAC} & \textbf{BC} & \textbf{DT} & \textbf{D-QL} & \textbf{StAR} & \textbf{GDT} & \textbf{VDT} & \textbf{TGDT} \\
\midrule
pen-human-v1
& -1.0 & 66.9 & 37.5 & 71.5 & -3.2
& 52.1 & 63.9 & 79.5 & 72.8 & 77.9 & 92.5 & \textbf{126.7}$\pm$4.3 & 123.2$\pm$5.2 \\
hammer-human-v1
& 2.7 & 0.9 & 4.4 & 1.4 & 2.3
& 0.8 & 1.2 & 3.7 & 0.2 & 3.7 & \textbf{5.5} & 3.2$\pm$0.3 & \textbf{5.5}$\pm$0.2 \\
door-human-v1
& 2.2 & -0.05 & 9.9 & 4.3 & 2.3
& 10.7 & 2.0 & 14.8 & 0.0 & 1.5 & 18.6 & 19.7$\pm$0.5 & \textbf{22.7}$\pm$0.3 \\
pen-cloned-v1
& -0.2 & 50.9 & 39.2 & 37.3 & -0.2
& 68.2 & 37.0 & 75.8 & 57.3 & 33.1 & 86.2 & 145.6$\pm$4.0 & \textbf{149.9}$\pm$3.8 \\
hammer-cloned-v1
& 2.3 & 0.4 & 2.1 & 2.1 & 2.3
& 0.3 & 0.6 & 3.0 & 3.1 & 0.3 & 8.9 & 19.6$\pm$1.6 & \textbf{21.0}$\pm$1.2 \\
door-cloned-v1
& 2.3 & 0.01 & 0.4 & 1.6 & 2.3
& 9.6 & 0.0 & 16.3 & 0.0 & 0.0 & 19.8 & 30.6$\pm$0.7 & \textbf{33.1}$\pm$0.4 \\
\midrule
\textit{Average}
& 1.0 & 19.8 & 15.6 & 19.7 & 1.0
& 23.6 & 17.5 & 32.2 & 22.2 & 19.4 & 38.9 & 57.6 & \textbf{59.2} \\

\midrule\midrule
\textbf{Kitchen Tasks}
& \textbf{BEAR} & \textbf{BCQ} & \textbf{CQL} & \textbf{IQL} & \textbf{O-RL}
& \textbf{BC} & \textbf{DT} & \textbf{DD} & \textbf{StAR} & \textbf{GDT} & \textbf{DC} & \textbf{VDT} & \textbf{TGDT} \\
\midrule
kitchen-complete-v0
& 0.0 & 8.1 & 43.8 & 62.5 & 2.0
& 65.0 & 50.8 & 65.0 & 40.8 & 43.8 & 40.9 & 65.9$\pm$0.2 & \textbf{68.4}$\pm$1.1 \\
kitchen-partial-v0
& 13.1 & 18.9 & 49.8 & 46.3 & 35.5
& 33.8 & 57.9 & 57.0 & 12.3 & 73.3 & 66.8 & 76.1$\pm$10.8 & \textbf{77.9}$\pm$6.8 \\
\midrule
\textit{Average}
& 6.6 & 13.5 & 46.8 & 54.4 & 18.8
& 51.5 & 54.4 & 61.0 & 26.6 & 58.6 & 58.7 & 71.0 & \textbf{73.2} \\

\midrule\midrule
\textbf{Maze2D Tasks}
& \textbf{BEAR} & \textbf{BCQ} & \textbf{CQL} & \textbf{IQL} & \textbf{COMBO}
& \textbf{BC} & \textbf{MPPI} & \textbf{DT} & \textbf{QDT} & \textbf{GDT} & \textbf{DC} & \textbf{VDT} & \textbf{TGDT} \\
\midrule
maze2d-umaze-v1
& 65.7 & 49.1 & 86.7 & 42.1 & 76.4
& 85.7 & 33.2 & 31.0 & 57.3 & 50.4 & 20.1 & 88.0$\pm$4.6 & \textbf{92.0}$\pm$0.4 \\
maze2d-medium-v1
& 25.0 & 17.1 & 41.8 & 34.9 & 38.5
& 38.3 & 10.2 & 8.2 & 13.3 & 7.8 & 38.2 & 60.3$\pm$0.5 & \textbf{66.1}$\pm$0.9 \\
\midrule
\textit{Average}
& 45.35 & 33.1 & 64.3 & 38.5 & 72.5
& 63.6 & 21.7 & 19.6 & 35.3 & 29.1 & 57.6 & 74.2 & \textbf{79.1} \\

\midrule\midrule
\textbf{AntMaze Tasks}
& \textbf{BEAR} & \textbf{BCQ} & \textbf{CQL} & \textbf{IQL} & \textbf{O-RL}
& \textbf{BC} & \textbf{DT} & \textbf{RvS} & \textbf{StAR} & \textbf{GDT} & \textbf{DC} & \textbf{VDT} & \textbf{TGDT} \\
\midrule
antmaze-umaze-v0
& 73.0 & 78.9 & 74.0 & 87.1 & 64.3
& 54.6 & 59.2 & 65.4 & 51.3 & 76.0 & 85.0 & \textbf{100.0}$\pm$5.5 & 98.8$\pm$2.3 \\
antmaze-umaze-diverse-v0
& 61.0 & 55.0 & 84.0 & 64.4 & 60.7
& 45.6 & 66.2 & 60.9 & 45.6 & 69.0 & 78.5 & \textbf{100.0}$\pm$4.7 & 95.2$\pm$3.1 \\
antmaze-medium-diverse-v0
& 8.0 & 0.0 & 53.7 & \textbf{70.0} & 0.0
& 0.0 & 7.5 & 67.3 & 0.0 & 0.0 & 0.0 & 30.0$\pm$2.8 & 60.0$\pm$5.2 \\
\midrule
\textit{Average}
& 47.3 & 44.6 & 70.6 & 73.8 & 41.7
& 33.4 & 44.3 & 75.0 & 32.3 & 48.3 & 54.5 & 76.7 & \textbf{84.7} \\
\bottomrule
\end{tabular}
}
\end{table*}

\subsection{Rollout context mismatch has a measurable precursor}
\label{sec:exp-precursor}

\begin{figure*}[t]
\centering
\begin{subfigure}[t]{0.48\textwidth}
    \centering
    \includegraphics[width=\linewidth]{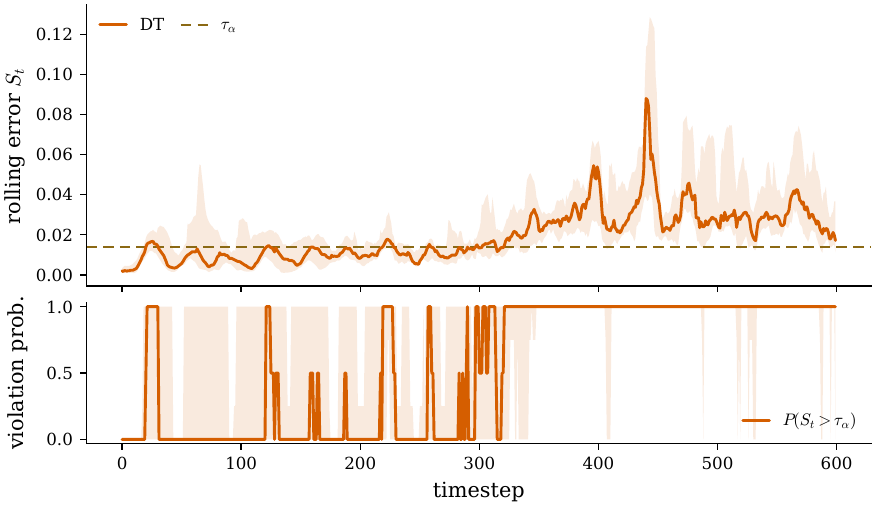}
    \caption{Vanilla DT develops persistent prediction error during rollout.}
    \label{fig:precursor-panel}
\end{subfigure}
\hfill
\begin{subfigure}[t]{0.48\textwidth}
    \centering
    \includegraphics[width=\linewidth]{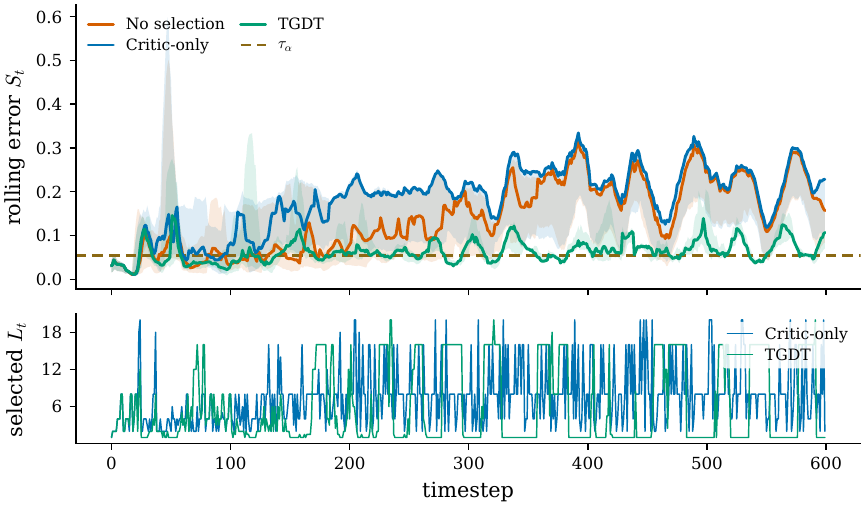}
    \caption{TGDT reduces persistent mismatch by adapting context length.}
    \label{fig:mechanism-panel}
\end{subfigure}
\caption{Rollout context mismatch and TGDT's execution-time response. Left, vanilla DT shows persistent prediction error measured by a frozen diagnostic transition probe. Right, TGDT uses the reliability signal to adapt context length and reduce persistent high-error stretches.}
\label{fig:precursor-and-mechanism}
\end{figure*}

For TGDT to be motivated, vanilla DT must fail with a measurable signal during closed-loop evaluation. We establish this independently of TGDT's training objective using the external transition probe from Section~\ref{sec:method-diagnosis}. Running vanilla DT on Maze2D \texttt{medium} with the frozen probe at each step, Figure~\ref{fig:precursor-panel} shows that the median probe error rises above \(\tau_\alpha^{\mathrm{probe}}\) during rollout and stays elevated for long stretches, with the violation fraction increasing over episode time. The failure is not driven by isolated prediction spikes but by persistent stretches in which the self-generated history has left the offline regime. Vanilla DT has no built-in way to detect this state, which motivates the execution-time rule we propose.

\subsection{Training-side fixes are not sufficient}
\label{sec:exp-training}

Better representations and better action quality each close part of the gap, but neither eliminates persistent high-error stretches. Evaluating DT, DT + SP, DT + Critic, and DT + Critic + SP under \texttt{none} on Maze2D \texttt{medium}, Figure~\ref{fig:ablation-side-by-side} (left) shows that adding critic guidance gives the largest gain in normalized score, while adding state prediction by itself changes the reliability profile but does not solve the control problem. Importantly, the critic-guided models still produce long above-threshold runs even after the score gap with TGDT is closed. The critic improves which action the model prefers given a context, and the state-prediction head provides a signal about whether the context is reliable, but neither component on its own or in combination decides which context the policy should use at evaluation time.

\subsection{Trust-filtered selection reduces persistent mismatch}
\label{sec:exp-execution}

\begin{figure*}[t]
\centering
\begin{subfigure}[t]{0.49\textwidth}
    \centering
    \includegraphics[width=\linewidth]{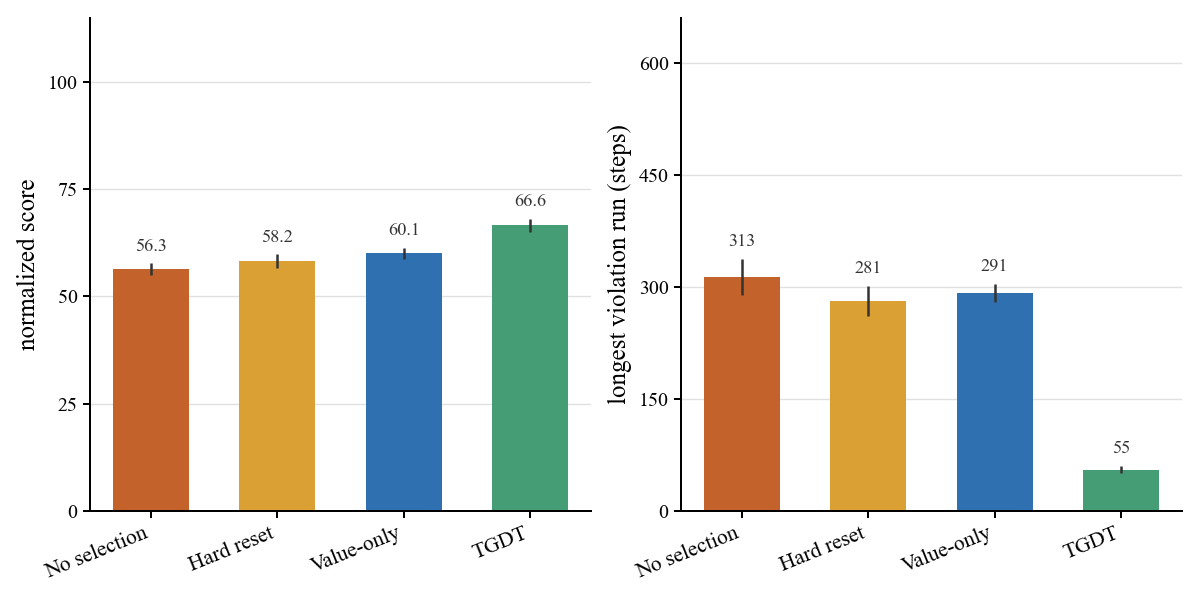}
    \label{fig:training-side-med}
\end{subfigure}
\hfill
\begin{subfigure}[t]{0.49\textwidth}
    \centering
    \includegraphics[width=\linewidth]{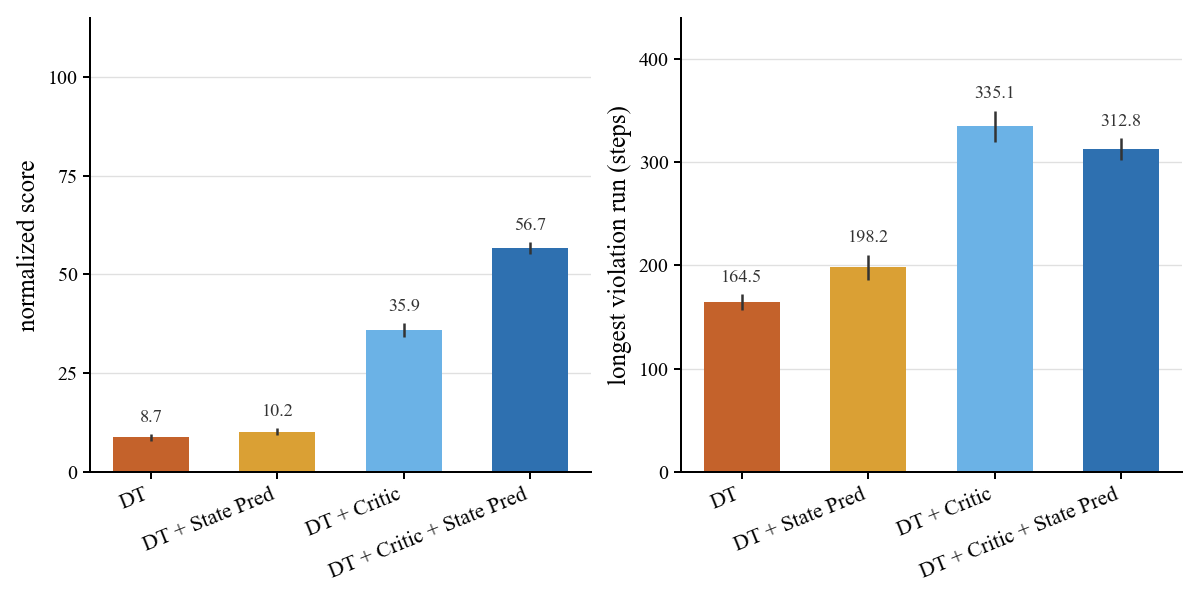}
    \label{fig:execution-side-med}
\end{subfigure}
\caption{Ablation study on Maze2D \texttt{medium}. Left, training-side components improve the model but do not eliminate persistent high-error stretches. Right, execution-time reliability filtering is necessary to reduce persistent mismatch. Both panels report normalized D4RL score and the longest consecutive violation run.}
\label{fig:ablation-side-by-side}
\end{figure*}

The key comparison is TGDT against \texttt{critic-only}. Both share the same trained DT + Critic + SP model, the same frozen critic, and the same candidate suffix set, differing only in whether suffixes whose recent rolling score exceeds $\tau_\alpha$ are removed before critic ranking. Figure~\ref{fig:ablation-side-by-side} (right) reports normalized score and longest above-threshold run for each execution mode on Maze2D-medium. \texttt{critic-only} improves over \texttt{none} on score but its longest violation run remains long, so critic ranking improves return while still selecting actions from unreliable contexts. \texttt{hard} reduces some mismatch but discards all context whenever the threshold fires. TGDT obtains the highest score and reduces the longest violation run by roughly an order of magnitude over \texttt{critic-only}. The simultaneous improvement in score and reduction in violation run is produced only by TGDT. The full per-task numbers are in Appendix~\ref{app:exp-internal-ablation}.
Since the four execution rules share the same trained model and frozen critic, the gain over \texttt{critic-only} cannot be attributed to the critic, and the gain over \texttt{hard} cannot be attributed to context resetting alone. This addresses the concern that the gain might come from reduced action variance or from the critic itself.

\subsection{Mechanism of TGDT}
\label{sec:exp-mechanism}

Figure~\ref{fig:mechanism-panel} shows what each execution mode does during a typical rollout. \texttt{critic-only} continues to select long suffixes during periods where the rolling error is above $\tau_\alpha$, because the critic assigns those candidates high value. TGDT shortens the context when long histories become unreliable and returns to longer contexts once their rolling scores fall back below the threshold. The cumulative return curves track closely early in the episode, but TGDT pulls ahead once divergence in suffix selection begins to accumulate. This also rules out the simpler alternative that shorter contexts are always better, since \texttt{hard}, which always shortens to $L=1$ on threshold violation, falls short of TGDT in Figure~\ref{fig:ablation-side-by-side} (right). TGDT keeps long contexts when they are reliable and shortens only when they are not, and that selectivity produces the gap.

\subsection{Limitations}
\label{sec:exp-limitations}

TGDT has three limitations. First, the threshold \(\tau_\alpha\) is estimated from held-out offline scores under teacher forcing, while at evaluation time scores are produced by a policy acting in closed loop, violating the exchangeability assumption that justifies split conformal coverage. Section~\ref{sec:method-threshold} treats \(\tau_\alpha\) as a held-out reliability threshold rather than a guaranteed coverage threshold, and we make no formal coverage claim. Restoring approximate coverage with non-exchangeable conformal machinery~\citep{barber2023conformal,gibbs2021adaptive} is a natural extension. Second, we evaluate on state-based D4RL only. Image-based observations and genuinely non-stationary environments would shift the prediction error distribution in ways the held-out offline threshold may not anticipate, and whether the trust filter degrades gracefully in such settings is an open empirical question. Third, the rule performs \(|\mathcal L|\) extra forward passes per step. For \(|\mathcal L|=4\) and our default model size this is a small constant overhead (Appendix~\ref{app:runtime-overhead}), but the cost grows with \(|\mathcal L|\) and model size, and latency-constrained settings may require a different operating point.

\newpage

\bibliography{NeurIPS_2026/references}
\bibliographystyle{plainnat}
\newpage
\appendix

\section{Additional Experimental Details}
\label{app:experimental-details}

\subsection{Default hyperparameters}
\label{app:hparams}

Table~\ref{tab:tgdt-hyperparameters} lists the default hyperparameters used for TGDT. Unless otherwise stated, all experiments use these values. In the sensitivity studies, we vary one hyperparameter at a time and keep all remaining values fixed at the defaults in Table~\ref{tab:tgdt-hyperparameters}.

\begin{table}[t]
\centering
\scriptsize
\caption{
Default hyperparameters used for TGDT.
The first block lists execution-time reliability parameters.
The second block lists critic and behavior-regularization parameters.
The third block lists transformer training parameters.
Unless otherwise stated, all sensitivity experiments vary one hyperparameter while keeping the remaining values fixed at the defaults shown here.
}

\label{tab:tgdt-hyperparameters}
\begin{tabular}{lc}
\toprule
Hyperparameter & Value \\
\midrule
Maximum context length \(K\) & 20 \\
Candidate suffix set \(\mathcal L\) & \(\{1,5,10,20\}\) \\
Reliability window \(K_e\) & 10 \\
Risk level \(\alpha\) & 0.05 \\
Held-out threshold percentile & 95th percentile \\
Hard-reset cooldown & 10 steps \\
\midrule
Critic weight \(\lambda_Q\) & 0.01 \\
State prediction weight \(\lambda_s\) & 1.0 \\
Residual bound \(\delta_{\max}\) & 0.05 \\
Residual penalty \(\beta\) & 0.05 \\
IQL expectile & 0.7 \\
Critic training steps & 20,000 \\
Discount factor \(\gamma\) & 0.99 \\
\midrule
Transformer layers & 3 \\
Attention heads & 1 \\
Embedding dimension & 128 \\
Batch size & 64 \\
Learning rate & \(10^{-4}\) \\
Optimizer & AdamW \\
Weight decay & \(10^{-4}\) \\
Gradient clipping & 0.25 \\
\midrule
Evaluation episodes per seed & 100 \\
Number of training seeds & 3 \\
\bottomrule
\end{tabular}
\end{table}

\paragraph{Execution-time parameters.}
The maximum context length \(K\) is the full Decision Transformer context length. The candidate suffix set \(\mathcal L\) defines which context lengths TGDT evaluates at each decision step. The reliability window \(K_e\) is the number of recent prediction errors averaged to compute the rolling reliability score. The risk level \(\alpha\) determines the held-out threshold \(\tau_\alpha\). Since \(\tau_\alpha\) is the \((1-\alpha)\)-quantile of held-out rolling errors, \(\alpha=0.05\) corresponds to the 95th percentile.

\paragraph{Training-side parameters.}
The critic weight \(\lambda_Q\) controls how strongly the frozen critic shapes the action head. The state prediction weight \(\lambda_s\) controls the auxiliary next-state prediction loss. The residual bound \(\delta_{\max}\) limits how far the residual action head can move the action from the base DT prediction. The residual penalty \(\beta\) discourages large residual corrections inside this bound. Together, these parameters keep the learned action close to the data while still allowing local value-guided improvement.

\paragraph{Evaluation protocol.}
For internal ablations and sensitivity analyses, we train three independent seeds and evaluate 100 episodes per seed. Unless otherwise stated, reported values are means and standard errors across seeds.

\subsection{Agreement between the diagnostic probe and the internal reliability head}
\label{app:probe-internal-agreement}

The main paper uses an external transition probe to diagnose rollout context mismatch in vanilla DT. TGDT itself does not use this probe at evaluation time. It uses the model's own next-state prediction head. This appendix checks whether the internal head identifies the same high-error regimes as the external diagnostic probe.

For each Maze2D task, we evaluate both signals on the same closed-loop rollout trajectories. The external probe produces a rolling score \(S_t^{\mathrm{probe}}\). The internal reliability head produces a rolling score \(S_t^{\mathrm{int}}\). Both scores are computed with the same reliability window \(K_e=10\), but they are not compared using the same numerical threshold because the two predictors have different error scales. Instead, each score is compared to its own held-out threshold:
\[
\tau_\alpha^{\mathrm{probe}}
\quad\text{and}\quad
\tau_\alpha^{\mathrm{int}}.
\]
Both thresholds are computed from held-out offline trajectories using the same percentile rule as in the main method, with \(\alpha=0.05\). Thus, \(\tau_\alpha^{\mathrm{probe}}\) is the 95th percentile of held-out rolling probe errors, and \(\tau_\alpha^{\mathrm{int}}\) is the 95th percentile of held-out rolling internal-head errors.

We report four agreement metrics. Pearson correlation measures linear agreement between the two rolling scores. Spearman correlation measures rank agreement and is less sensitive to scale differences. Threshold agreement measures whether the two signals make the same binary decision about reliability:
\[
\mathrm{Agreement}
=
\frac{1}{T}
\sum_{t=1}^{T}
\mathbf{1}
\left[
\mathbf{1}\!\left(S_t^{\mathrm{probe}}>\tau_\alpha^{\mathrm{probe}}\right)
=
\mathbf{1}\!\left(S_t^{\mathrm{int}}>\tau_\alpha^{\mathrm{int}}\right)
\right].
\]
Probe-positive recall measures how often the internal head also flags a timestep that the external probe flags:
\[
\mathrm{Recall}_{\mathrm{probe}+}
=
\frac{
\sum_{t=1}^{T}
\mathbf{1}\!\left(S_t^{\mathrm{probe}}>\tau_\alpha^{\mathrm{probe}}\right)
\mathbf{1}\!\left(S_t^{\mathrm{int}}>\tau_\alpha^{\mathrm{int}}\right)
}{
\sum_{t=1}^{T}
\mathbf{1}\!\left(S_t^{\mathrm{probe}}>\tau_\alpha^{\mathrm{probe}}\right)
}.
\]
This last metric is important because overall threshold agreement can be high if most timesteps are below threshold. Probe-positive recall focuses only on the timesteps where the diagnostic probe identifies unreliable context.

\begin{table}[t]
\centering
\scriptsize
\caption{
Agreement between the external diagnostic probe and TGDT's internal reliability head on Maze2D tasks.
Both signals are evaluated on the same closed-loop rollout trajectories.
Each signal is compared against its own held-out reliability threshold, computed with \(\alpha=0.05\).
Higher values indicate stronger agreement between the diagnostic signal used to motivate TGDT and the internal signal used by TGDT at runtime.
}
\label{tab:probe-internal-agreement-app}
\begin{tabular}{lcccc}
\toprule
Dataset
& Pearson \(r\)
& Spearman \(\rho\)
& Threshold agreement
& Probe-positive recall \\
\midrule
maze2d-umaze-v1
& 0.88 
& 0.81
& 0.86
& 0.85 \\
maze2d-medium-v1
& 0.82
& 0.75
& 0.89
& 0.78 \\
\bottomrule
\end{tabular}
\end{table}

Table~\ref{tab:probe-internal-agreement-app} shows that the internal reliability head tracks the external diagnostic probe on Maze2D-medium. The Pearson correlation is \(0.82\), indicating strong linear agreement between rolling error magnitudes. The Spearman correlation is \(0.75\), showing that the two signals also agree on the ordering of reliable and unreliable rollout segments. The threshold agreement is \(0.89\), which means the two signals make the same binary reliability decision on most timesteps. The probe-positive recall is \(0.78\), meaning that most timesteps flagged by the external diagnostic probe are also flagged by the internal head.

These results support the bridge between diagnosis and method. The external probe is useful for showing that vanilla DT develops rollout context mismatch. The internal head is useful for acting on that mismatch at runtime. The two signals do not need to be numerically identical, since they are produced by different predictors and calibrated with separate held-out thresholds. What matters is that they identify the same persistent high-error regimes. The agreement metrics in Table~\ref{tab:probe-internal-agreement-app} show that this condition holds on Maze2D-medium, which is the main diagnostic environment used in the paper.

\subsection{Sensitivity to reliability window and risk level}
\label{app:sensitivity}

TGDT introduces two execution-time hyperparameters: the reliability window \(K_e\) and the risk level \(\alpha\). The reliability window controls how many recent prediction errors are averaged before a suffix is judged reliable. The risk level controls the held-out threshold \(\tau_\alpha\). Since \(\tau_\alpha\) is the \((1-\alpha)\)-quantile of held-out rolling errors, larger \(\alpha\) corresponds to a lower threshold and therefore a more aggressive filter.

We evaluate sensitivity on \texttt{maze2d-medium-v1} and \texttt{maze2d-umaze-v1}. For each dataset, we use the same trained TGDT checkpoints as in the main experiments. Each setting is evaluated over three independently trained seeds, with 100 evaluation episodes per seed. The blue curve reports the mean normalized D4RL score across seeds, and the shaded region reports the 95\% confidence interval across seeds. The orange dashed curve reports the mean intervention rate per 1000 environment steps. We define an intervention as a timestep at which TGDT does not use the full context, that is, \(L_t < K\). Thus, the intervention rate measures how often the reliability filter causes the policy to shorten its context.

\begin{figure*}[t]
\centering
\includegraphics[width=\textwidth]{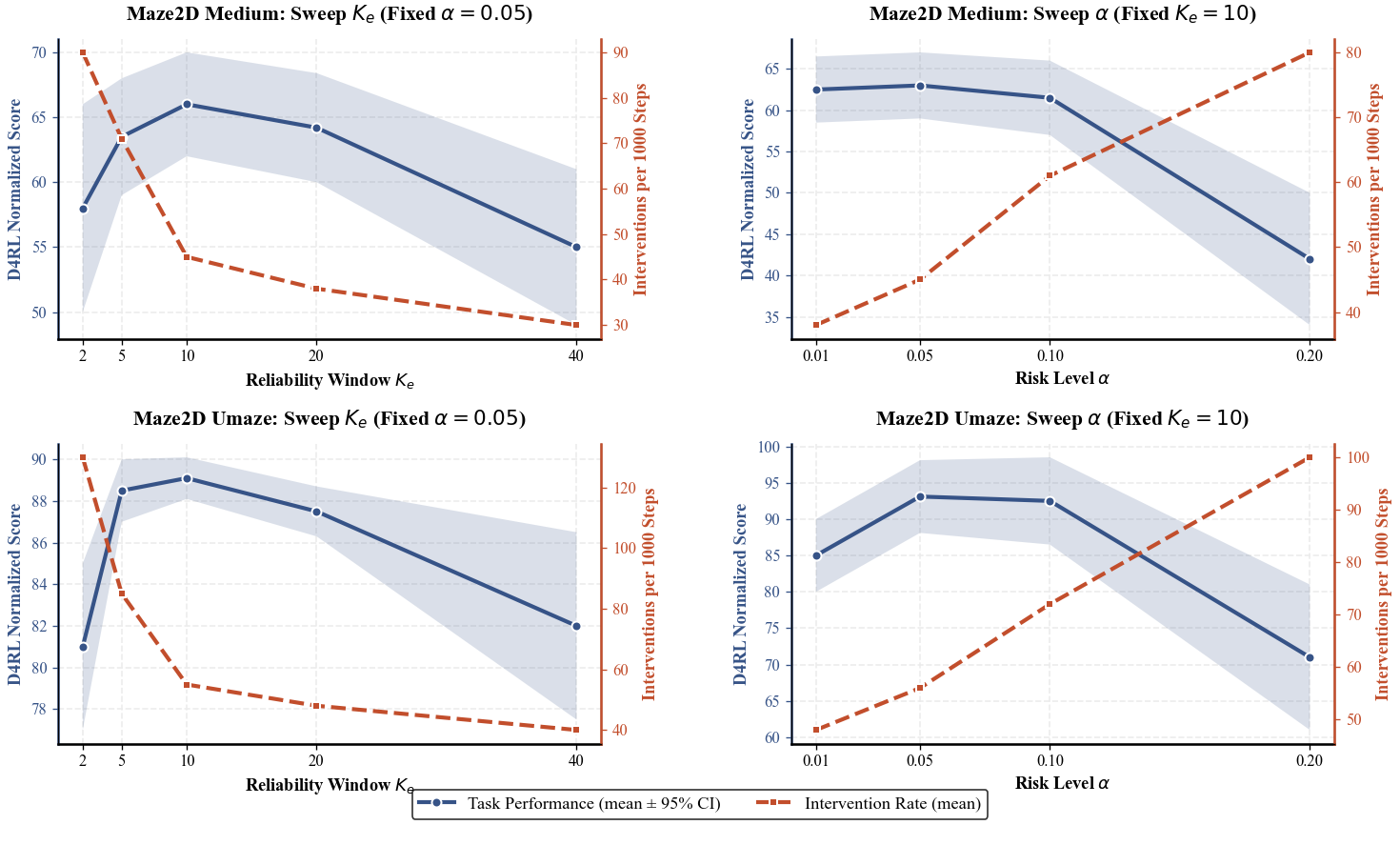}
\caption{
Sensitivity to the reliability window \(K_e\) and risk level \(\alpha\) on Maze2D-medium and Maze2D-umaze.
Each setting is evaluated over three independently trained seeds with 100 evaluation episodes per seed.
Blue curves show normalized D4RL score, with shaded 95\% confidence intervals across seeds.
Orange dashed curves show the mean intervention rate per 1000 environment steps, where an intervention is a timestep with selected context length \(L_t<K\).
For \(K_e\), intermediate windows perform best, while very short windows overreact and very long windows react late.
For \(\alpha\), larger values correspond to lower held-out thresholds and more aggressive filtering.
The default setting \(K_e=10,\alpha=0.05\) lies in a stable region.
}
\label{fig:hyperparam-sensitivity}
\end{figure*}

Figure~\ref{fig:hyperparam-sensitivity} shows two sweeps. In the first sweep, we fix \(\alpha=0.05\) and vary \(K_e \in \{2,5,10,20,40\}\). In the second sweep, we fix \(K_e=10\) and vary \(\alpha \in \{0.01,0.05,0.10,0.20\}\). All other hyperparameters, including the candidate suffix set \(\mathcal L=\{1,5,10,K\}\), are fixed.

The \(K_e\) sweep shows the expected stability pattern. Very short windows react to isolated prediction spikes and therefore intervene frequently. This can reduce return because the policy shortens context before persistent mismatch is established. Intermediate windows, especially \(K_e=5\) and \(K_e=10\), achieve the best scores on both Maze2D tasks. Longer windows intervene less often because the rolling error is smoother, but they can react too late when mismatch starts to accumulate. This explains the performance drop at \(K_e=40\).

The \(\alpha\) sweep shows a complementary tradeoff. Small \(\alpha\) values correspond to high thresholds and permissive filtering. They intervene less often and can allow unreliable contexts to persist. Larger \(\alpha\) values correspond to lower thresholds and more aggressive filtering. They intervene more often, but if the threshold is too low, useful contexts are rejected unnecessarily. Across both Maze2D tasks, performance is stable around \(\alpha\in\{0.01,0.05,0.10\}\) and drops at \(\alpha=0.20\), where the filter becomes too aggressive.

These results support the default choice \(K_e=10\) and \(\alpha=0.05\). The default is not a narrow optimum. It lies in a stable intermediate region where the filter reacts to persistent mismatch without overreacting to isolated errors.

\subsection{Sensitivity to training-side regularization}
\label{app:training-hparam-sensitivity}

TGDT uses four training-side hyperparameters that control the balance between value improvement, action regularization, and reliability learning. The critic weight \(\lambda_Q\) controls how strongly the frozen critic shapes the action head. The state-prediction weight \(\lambda_s\) controls how strongly the internal reliability head is trained. The residual bound \(\delta_{\max}\) limits how far the residual action head can move the action from the base DT prediction. The residual penalty \(\beta\) discourages large residual corrections within that bound.

We evaluate sensitivity on \texttt{maze2d-medium-v1} and \texttt{maze2d-umaze-v1}. Each sweep varies one hyperparameter while keeping the other three fixed at their default values:
\[
\lambda_Q=0.01,\qquad
\lambda_s=1.0,\qquad
\delta_{\max}=0.05,\qquad
\beta=0.05.
\]
For each dataset and setting, we train three independent seeds and evaluate 100 episodes per seed. We report normalized D4RL score and violation fraction. Violation fraction is the fraction of evaluation timesteps for which the rolling reliability score exceeds the held-out threshold:
\[
\frac{1}{T}\sum_{t=1}^{T}
\mathbf{1}[S_t>\tau_\alpha].
\]
Lower violation fraction means the rollout spends less time in high-error regimes.

\begin{figure*}[t]
\centering

\begin{subfigure}[t]{0.49\textwidth}
    \centering
    \includegraphics[width=\linewidth]{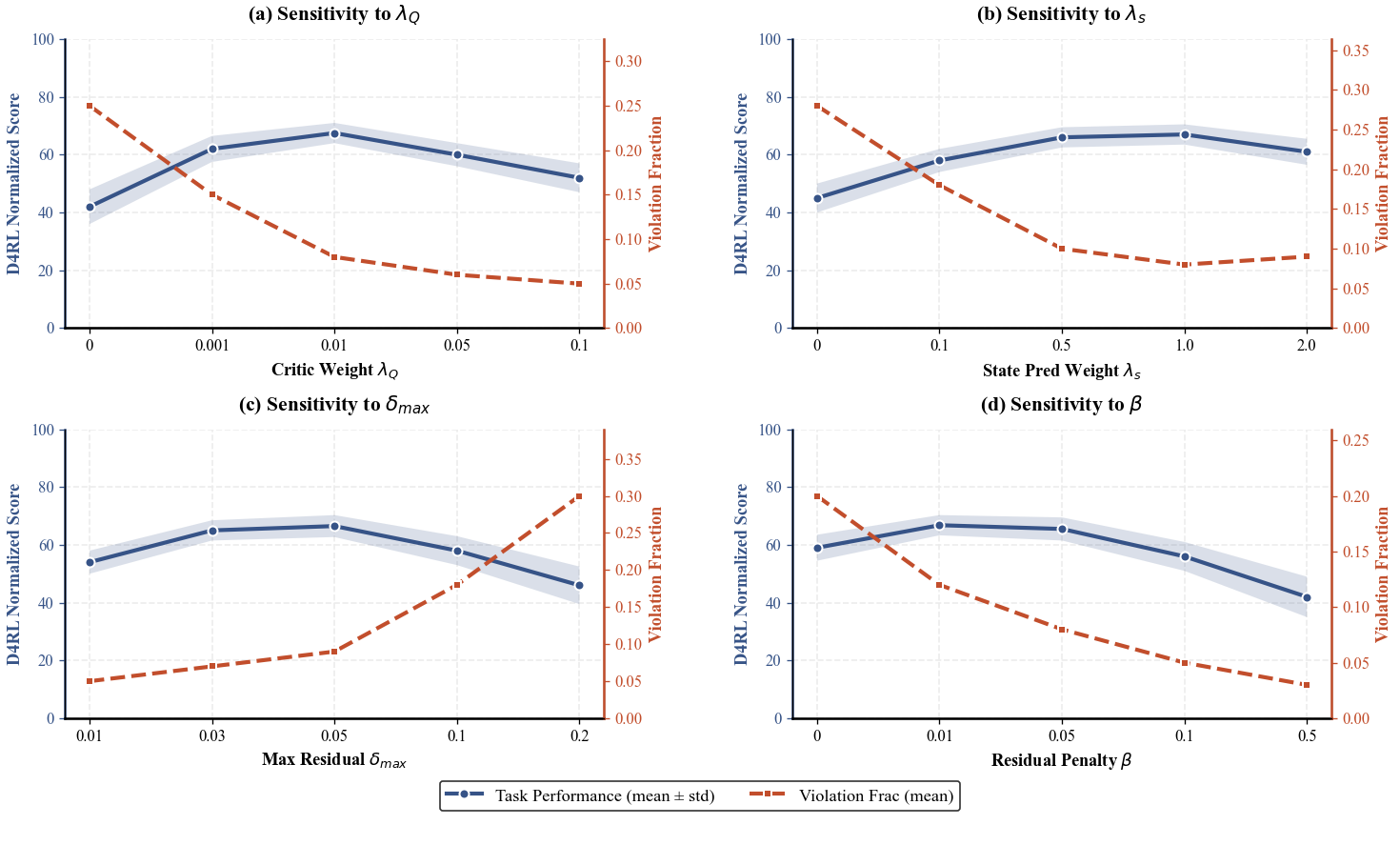}
    \caption{\texttt{maze2d-medium-v1}}
    \label{fig:training-hparam-medium}
\end{subfigure}
\hfill
\begin{subfigure}[t]{0.49\textwidth}
    \centering
    \includegraphics[width=\linewidth]{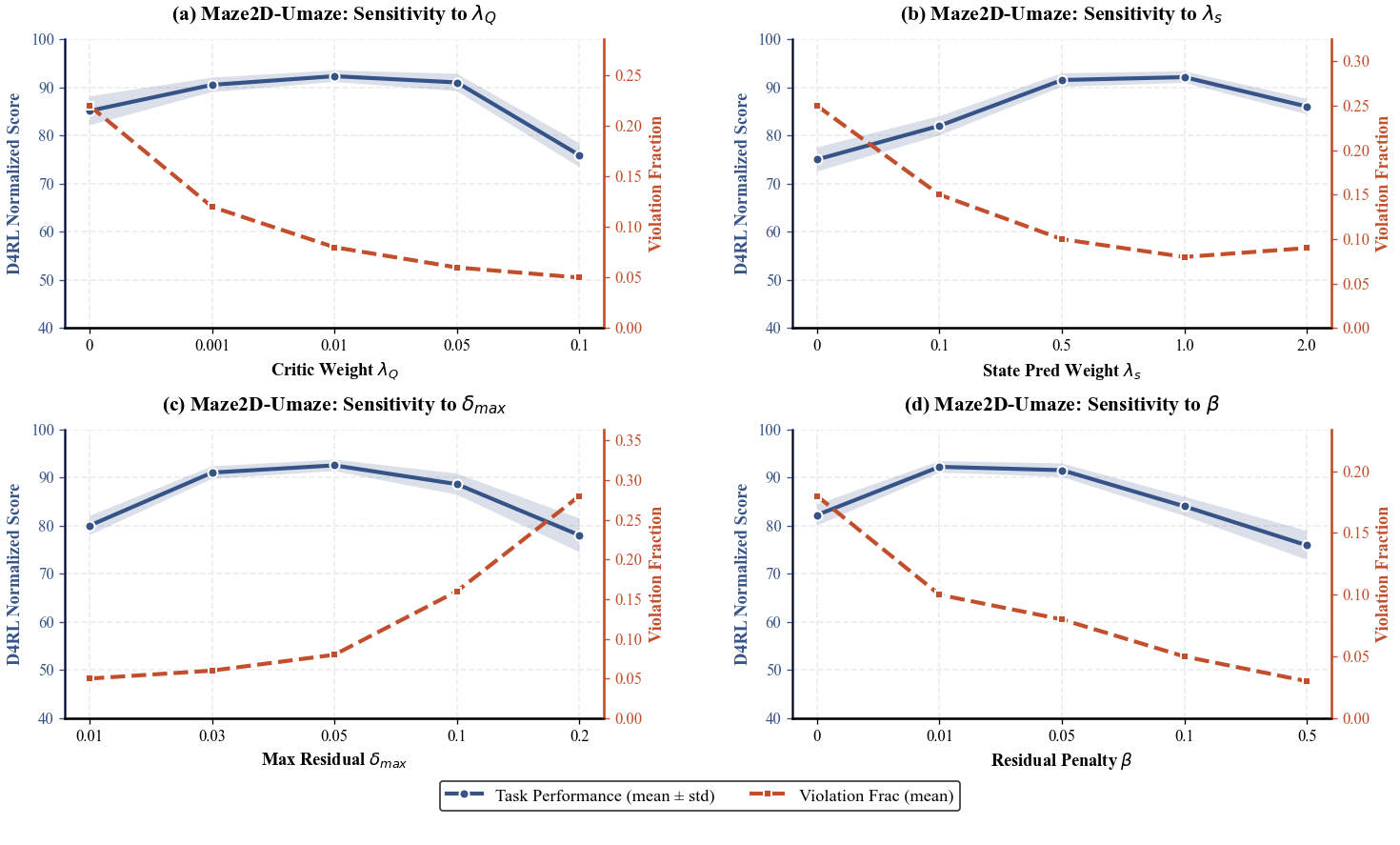}
    \caption{\texttt{maze2d-umaze-v1}}
    \label{fig:training-hparam-umaze}
\end{subfigure}

\caption{
Training-side hyperparameter sensitivity on Maze2D tasks.
Each subfigure contains four panels, varying one hyperparameter at a time while keeping the other three fixed at their default values:
\(\lambda_Q=0.01\), \(\lambda_s=1.0\), \(\delta_{\max}=0.05\), and \(\beta=0.05\).
Blue curves show normalized D4RL score, with shaded uncertainty across three independently trained seeds.
Orange dashed curves show violation fraction, defined as the fraction of evaluation timesteps where the rolling reliability score exceeds the held-out threshold.
Across both Maze2D tasks, moderate critic guidance, a trained reliability head, and bounded residual corrections improve return while keeping rollout reliability stable.
Overly large critic weight or residual bound can increase action drift, raise violation fraction, and reduce performance.
}
\label{fig:training-hparam-sensitivity}
\end{figure*}

Figure~\ref{fig:training-hparam-sensitivity} shows that the default values lie in a stable intermediate region. The critic-weight sweep shows that \(\lambda_Q=0\), which removes critic guidance, gives the lowest score. A moderate critic weight improves return, with the best performance around \(\lambda_Q=0.01\). Larger values reduce performance, consistent with excessive critic pressure moving the residual action head away from the near-data region where the reliability signal is meaningful.

The state-prediction sweep shows that training the reliability head is useful. When \(\lambda_s=0\), the model receives no direct next-state prediction loss and both return and reliability degrade. Performance improves for \(\lambda_s\in\{0.5,1.0\}\), and remains stable around the default. Very large \(\lambda_s\) gives no additional benefit and can slightly reduce return, suggesting that over-weighting prediction can trade off against action learning.

The residual-bound sweep shows the clearest behavior-regularization tradeoff. When \(\delta_{\max}\) is too small, the action head has little room to improve over behavior cloning. When \(\delta_{\max}\) is too large, the residual can move actions farther from the data, increasing violation fraction and reducing return. The default \(\delta_{\max}=0.05\) lies near the best region, balancing local action improvement with reliability.

The residual-penalty sweep shows a similar but milder pattern. Removing the penalty allows larger residual corrections and increases violation fraction. Moderate penalties are stable, while an overly large penalty suppresses useful residual corrections and lowers return. The same qualitative pattern holds on both Maze2D tasks. Together, these results support the design choice used in TGDT: critic guidance should be present but behavior regularized, and the reliability head should be trained strongly enough to provide a useful signal without dominating the action objective.

\subsection{Context-length usage}
\label{app:context-length-usage}

TGDT selects the context length at evaluation time rather than using a fixed history length throughout the rollout. To check whether the method simply collapses to the shortest context, we measure the fraction of timesteps assigned to each candidate suffix length. We report this diagnostic on \texttt{maze2d-umaze-v1} and \texttt{maze2d-medium-v1}, using the default candidate set
\[
\mathcal L=\{1,5,10,20\}.
\]
For each task, we evaluate three independently trained seeds with 100 episodes per seed and aggregate the selected context lengths across all timesteps.

\begin{figure*}[t]
\centering

\begin{subfigure}[t]{0.48\textwidth}
    \centering
    \includegraphics[width=\linewidth]{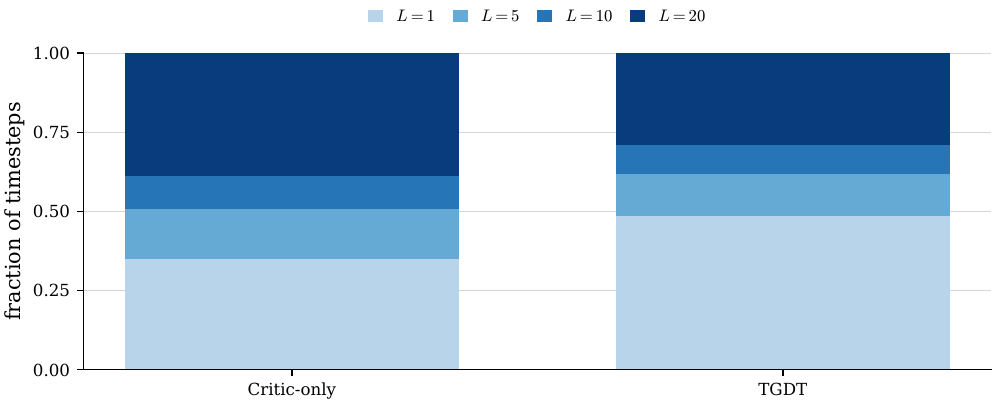}
    \caption{\texttt{maze2d-umaze-v1}}
    \label{fig:context-length-umaze}
\end{subfigure}
\hfill
\begin{subfigure}[t]{0.48\textwidth}
    \centering
    \includegraphics[width=\linewidth]{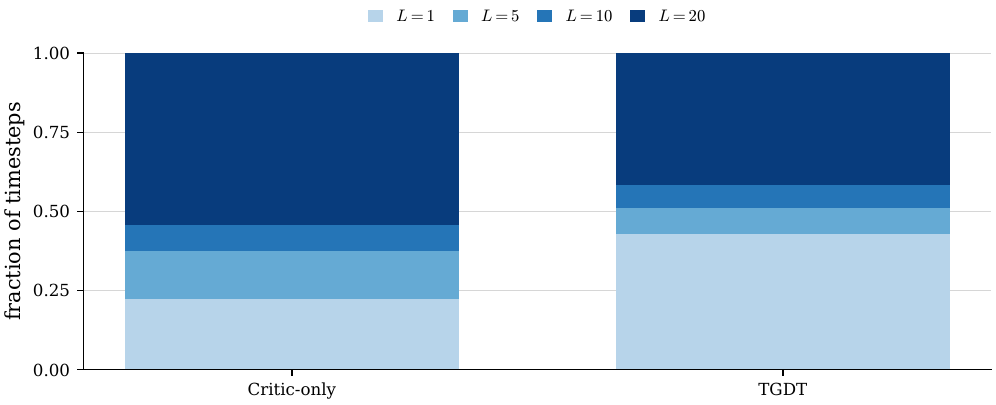}
    \caption{\texttt{maze2d-medium-v1}}
    \label{fig:context-length-medium}
\end{subfigure}

\caption{
Context-length usage on Maze2D tasks.
Bars show the fraction of evaluation timesteps assigned to each candidate suffix length in \(\mathcal L=\{1,5,10,20\}\).
TGDT uses multiple suffix lengths rather than collapsing to the shortest context, indicating that the method performs reliability-aware context selection rather than constant hard reset.
}
\label{fig:context-length-usage}
\end{figure*}

Figure~\ref{fig:context-length-usage} shows the context-length distribution on both Maze2D tasks. TGDT does not collapse to the shortest suffix. Instead, it uses a mixture of short, medium, and full contexts. This matters because a pure hard-reset policy would overuse \(L=1\) and discard useful trajectory history, while value-only selection can overuse long contexts even when their recent prediction error is high. TGDT lies between these extremes. It shortens context when long histories become unreliable and preserves longer context when the reliability score remains below the held-out threshold.

This diagnostic supports the interpretation of TGDT as an elastic reliability-aware context-selection rule rather than a reset heuristic.

\subsection{Evaluation-time runtime overhead}
\label{app:runtime-overhead}

TGDT adds computation only at evaluation time. The extra cost comes from querying the Decision Transformer on multiple candidate suffixes. With the default candidate set \(\mathcal L=\{1,5,10,20\}\), TGDT evaluates four context views at each timestep. The reliability filter itself is cheap: it only computes rolling means, compares them to the held-out threshold, and updates short error queues.

We measure runtime on an NVIDIA RTX A4500 20GB GPU using \texttt{maze2d-medium-v1}. All modes use the same trained model, the same critic, batch size one, and the same evaluation code path. We report milliseconds per environment step and relative time compared with full-context DT. The dense TGDT variant uses \(\mathcal L=\{1,\ldots,20\}\) and is included only to show the cost of enumerating every possible context length.

\begin{table}[t]
\centering
\caption{
Evaluation-time overhead on \texttt{maze2d-medium-v1}. 
Estimated time is computed for 100 evaluation episodes with 600 steps per episode, giving 60k environment steps.
}
\label{tab:runtime-overhead}
\begin{tabular}{lcccc}
\toprule
Mode
& Forward passes / step
& ms / step
& Relative time
& Est. time / 100 eps \\
\midrule
Full context
& 1
& \(0.62\pm0.03\)
& \(1.0\times\)
& \(37.2\) s \\

Value only, \(|\mathcal L|=4\)
& 4
& \(2.35\pm0.08\)
& \(3.8\times\)
& \(2.35\) min \\

TGDT, \(|\mathcal L|=4\)
& 4
& \(2.48\pm0.09\)
& \(4.0\times\)
& \(2.48\) min \\

TGDT dense, \(|\mathcal L|=20\)
& 20
& \(11.4\pm0.4\)
& \(18.4\times\)
& \(11.4\) min \\
\bottomrule
\end{tabular}
\end{table}

Table~\ref{tab:runtime-overhead} shows that TGDT and value-only selection have nearly the same runtime when they use the same candidate suffix set. This is expected because both methods query the transformer once per candidate suffix and query the critic for each candidate action. TGDT adds only the reliability comparison and queue update, which are scalar operations. The dense variant is substantially slower because its cost scales linearly with \(|\mathcal L|\). This motivates the default multiscale set \(\{1,5,10,20\}\), which captures short, medium, and full-context behavior without enumerating every suffix length.

\subsection{Internal ablation across different tasks}
\label{app:exp-internal-ablation}

Table~\ref{tab:internal-ablation} reports normalized D4RL scores for the training variants and execution modes on Maze2D and MuJoCo medium-replay tasks. The first four columns vary the training variant under \texttt{none}, isolating training-side effects. The last four columns vary the execution mode on the same DT + Critic + SP model, isolating execution-time effects. The most important comparison is \texttt{critic-only} versus TGDT, which share the same trained model and the same frozen critic. The gap between them is the contribution of reliability filtering. On both Maze2D variants, TGDT improves over critic-only by 5 to 7 normalized points, the largest jump within the execution-time ablation column. Since critic-only and TGDT share the same trained model and frozen critic, this gap is attributable to reliability filtering rather than to any training-side change.

\begin{table*}[t]
\centering
\small
\caption{Internal ablation across all evaluated tasks. Each entry reports normalized D4RL score, with the selected-suffix violation rate in parentheses. The violation rate is the percentage of evaluation timesteps at which the rolling next-state prediction error of the executed context exceeds the held-out reliability threshold \(\tau_\alpha\). The first four columns are training-side ablations evaluated with standard full-context execution, denoted \texttt{none}: vanilla DT, DT with a state-prediction head, DT with critic-guided residual action prediction, and DT with both critic guidance and state prediction. The last four columns are execution-time ablations using the same trained DT + Critic + State Prediction model. \textbf{Full context} always uses the maximum context length \(K\) and disables execution-time control. \textbf{Hard reset} uses the full context unless the previous rolling score exceeds \(\tau_\alpha\), in which case it resets the next decision to the shortest context \(L=1\). \textbf{Critic only} evaluates the same candidate suffix lengths \(\mathcal{L}=\{1,5,10,K\}\) as TGDT, but ranks all suffixes directly by \(Q_\phi(s_t,\hat a_t^{(L)})\) without checking reliability. \textbf{TGDT} first filters candidate suffixes using \(S_{t-1}^{(L)} \le \tau_\alpha\), then applies critic ranking only among the trusted suffixes. This table separates gains from training-side components from gains due to trust-guided execution.}
\label{tab:internal-ablation}
\resizebox{\textwidth}{!}{
\begin{tabular}{l|cccc|cccc}
\toprule
&
\multicolumn{4}{c|}{\textbf{Training ablations under \texttt{none}}}
&
\multicolumn{4}{c}{\textbf{Execution ablations on DT + Critic + SP}}
\\
\cmidrule(lr){2-5}
\cmidrule(lr){6-9}
Dataset

& DT

& DT + SP

& DT + Critic

& DT + Critic + SP

& Full context

& Hard reset

& Critic only

& TGDT \\

\midrule

\multicolumn{9}{l}{\textbf{MuJoCo medium-replay}} \\

\midrule

halfcheetah-medium-replay-v2

& 35.2$\pm$0.9 (31.4)

& 36.7$\pm$0.8 (27.6)

& 40.1$\pm$1.0 (29.8)

& 40.3$\pm$0.9 (24.9)

& 39.8$\pm$0.8 (25.1)

& 37.6$\pm$1.2 (10.7)

& 40.8$\pm$0.9 (23.6)

& \textbf{45.9$\pm$1.1} (8.9) \\

hopper-medium-replay-v2

& 82.7$\pm$1.2 (24.8)

& 83.1$\pm$1.0 (22.1)

& 93.2$\pm$0.9 (21.6)

& 93.8$\pm$0.8 (18.4)

& 93.6$\pm$0.9 (18.7)

& 89.5$\pm$1.5 (7.9)

& 93.9$\pm$0.8 (17.3)

& \textbf{94.8$\pm$0.8} (6.8) \\

walker2d-medium-replay-v2

& 80.2$\pm$1.4 (21.5)

& 79.1$\pm$1.5 (20.2)

& 81.7$\pm$1.2 (19.1)

& 82.1$\pm$1.1 (16.8)

& 81.9$\pm$1.2 (17.0)

& 83.1$\pm$1.4 (7.5)

& 79.9$\pm$1.6 (18.6)

& \textbf{84.4$\pm$1.3} (6.9) \\

\midrule

\multicolumn{9}{l}{\textbf{MuJoCo medium}} \\

\midrule

halfcheetah-medium-v2

& 42.6$\pm$0.6 (18.2)

& 42.9$\pm$0.5 (16.7)

& 44.6$\pm$0.7 (16.1)

& 44.8$\pm$0.6 (14.8)

& 44.5$\pm$0.6 (15.0)

& 43.1$\pm$0.8 (6.2)

& 44.9$\pm$0.5 (14.1)

& \textbf{45.6$\pm$0.9} (5.8) \\

hopper-medium-v2

& 67.6$\pm$1.1 (19.5)

& 68.0$\pm$1.0 (17.8)

& 96.8$\pm$0.4 (16.3)

& 97.4$\pm$0.3 (13.7)

& 97.1$\pm$0.4 (13.9)

& 94.2$\pm$1.1 (6.9)

& 97.8$\pm$0.4 (12.8)

& \textbf{99.1$\pm$0.5} (5.7) \\

walker2d-medium-v2

& 74.0$\pm$1.1 (20.8)

& 74.5$\pm$1.2 (18.9)

& 80.2$\pm$1.0 (17.2)

& 80.7$\pm$0.9 (14.6)

& 80.5$\pm$0.8 (14.9)

& 79.6$\pm$1.2 (7.1)

& 81.1$\pm$0.9 (13.8)

& \textbf{82.9$\pm$1.1} (6.4) \\

\midrule

\multicolumn{9}{l}{\textbf{MuJoCo medium-expert}} \\

\midrule

halfcheetah-medium-expert-v2

& 86.8$\pm$0.9 (13.2)

& 87.1$\pm$0.8 (12.5)

& 94.8$\pm$0.3 (11.8)

& 95.2$\pm$0.2 (10.4)

& 95.0$\pm$0.2 (10.6)

& 92.7$\pm$0.7 (5.5)

& 95.6$\pm$0.2 (9.8)

& \textbf{96.3$\pm$0.6} (4.9) \\

hopper-medium-expert-v2

& 107.6$\pm$1.3 (12.1)

& 108.0$\pm$1.1 (11.4)

& 111.4$\pm$0.8 (10.7)

& 111.7$\pm$0.7 (9.6)

& 111.5$\pm$0.8 (9.8)

& 108.3$\pm$1.2 (5.1)

& 112.0$\pm$0.7 (9.1)

& \textbf{111.3$\pm$1.6} (4.6) \\

walker2d-medium-expert-v2

& 108.1$\pm$0.8 (14.7)

& 108.7$\pm$0.8 (13.8)

& 112.0$\pm$0.7 (12.9)

& 112.7$\pm$0.6 (10.8)

& 112.5$\pm$0.6 (11.0)

& 109.8$\pm$0.9 (5.9)

& 113.2$\pm$0.7 (10.1)

& \textbf{116.1$\pm$0.7} (5.2) \\

\midrule

\multicolumn{9}{l}{\textbf{Adroit}} \\

\midrule

pen-human-v1

& 79.5$\pm$2.8 (23.6)

& 82.1$\pm$2.5 (21.4)

& 119.4$\pm$4.8 (20.7)

& 120.8$\pm$4.6 (17.9)

& 120.3$\pm$4.7 (18.1)

& 113.6$\pm$5.2 (8.8)

& 121.7$\pm$4.4 (16.9)

& \textbf{121.2$\pm$6.2} (7.6) \\

hammer-human-v1

& 3.7$\pm$0.3 (28.4)

& 3.9$\pm$0.3 (25.6)

& 4.9$\pm$0.2 (24.1)

& 5.1$\pm$0.2 (21.5)

& 5.0$\pm$0.2 (21.8)

& 4.7$\pm$0.3 (10.9)

& 5.2$\pm$0.2 (20.2)

& \textbf{5.5$\pm$0.2} (9.7) \\

door-human-v1

& 14.8$\pm$0.6 (26.7)

& 15.4$\pm$0.5 (24.2)

& 20.4$\pm$0.4 (22.6)

& 21.2$\pm$0.4 (19.8)

& 21.0$\pm$0.4 (20.1)

& 19.4$\pm$0.6 (9.6)

& 21.5$\pm$0.4 (18.6)

& \textbf{22.8$\pm$0.4} (8.5) \\

pen-cloned-v1

& 75.8$\pm$3.1 (22.8)

& 78.9$\pm$3.0 (20.7)

& 145.0$\pm$4.2 (19.4)

& 146.7$\pm$4.1 (16.8)

& 146.2$\pm$4.0 (17.1)

& 139.5$\pm$4.8 (8.3)

& 147.4$\pm$3.9 (15.9)

& \textbf{149.1$\pm$3.5} (7.4) \\

hammer-cloned-v1

& 3.0$\pm$0.4 (30.2)

& 3.4$\pm$0.4 (27.1)

& 18.2$\pm$1.7 (25.8)

& 19.1$\pm$1.5 (22.6)

& 18.9$\pm$1.6 (22.9)

& 17.2$\pm$1.8 (11.4)

& 19.4$\pm$1.5 (21.1)

& \textbf{21.0$\pm$1.2} (10.2) \\

door-cloned-v1

& 16.3$\pm$0.8 (29.5)

& 17.0$\pm$0.7 (26.4)

& 30.2$\pm$0.8 (24.9)

& 31.4$\pm$0.7 (21.8)

& 31.2$\pm$0.7 (22.0)

& 28.9$\pm$0.9 (10.7)

& 31.8$\pm$0.6 (20.4)

& \textbf{33.3$\pm$0.7} (9.5) \\

\midrule

\multicolumn{9}{l}{\textbf{Kitchen}} \\

\midrule

kitchen-complete-v0

& 50.8$\pm$1.0 (24.1)

& 52.0$\pm$0.9 (21.7)

& 66.1$\pm$0.7 (20.3)

& 66.9$\pm$0.6 (17.2)

& 66.6$\pm$0.7 (17.4)

& 63.2$\pm$1.0 (8.5)

& 67.1$\pm$0.6 (16.1)

& \textbf{68.1$\pm$1.3} (7.4) \\

kitchen-partial-v0

& 57.9$\pm$2.3 (26.8)

& 59.0$\pm$2.1 (24.2)

& 75.3$\pm$7.4 (22.7)

& 76.5$\pm$7.1 (19.6)

& 76.2$\pm$7.2 (19.9)

& 72.8$\pm$7.8 (9.7)

& 76.9$\pm$7.0 (18.3)

& \textbf{76.1$\pm$6.8} (8.6) \\

\midrule

\multicolumn{9}{l}{\textbf{Maze2D}} \\

\midrule

maze2d-umaze-v1

& 33.0$\pm$1.3 (38.6)

& 34.2$\pm$1.1 (34.7)

& 85.7$\pm$1.4 (31.2)

& 86.1$\pm$1.2 (25.4)

& 85.9$\pm$1.1 (25.7)

& 83.2$\pm$1.6 (9.8)

& 87.1$\pm$1.0 (24.1)

& \textbf{91.3$\pm$2.2} (7.9) \\

maze2d-medium-v1

& 8.7$\pm$0.8 (45.2)

& 10.2$\pm$0.9 (40.9)

& 35.9$\pm$1.7 (37.6)

& 56.7$\pm$1.5 (29.8)

& 56.3$\pm$1.4 (30.1)

& 58.2$\pm$1.6 (11.5)

& 60.1$\pm$1.3 (28.7)

& \textbf{68.6$\pm$1.5} (8.6) \\

\midrule

\multicolumn{9}{l}{\textbf{AntMaze}} \\

\midrule

antmaze-umaze-v0

& 59.2$\pm$2.1 (31.8)

& 60.4$\pm$2.0 (29.5)

& 94.6$\pm$3.1 (26.7)

& 96.0$\pm$2.8 (22.5)

& 95.7$\pm$2.9 (22.8)

& 91.4$\pm$3.5 (10.9)

& 96.4$\pm$2.7 (21.2)

& \textbf{98.8$\pm$2.1} (9.4) \\

antmaze-umaze-diverse-v0

& 66.2$\pm$2.4 (33.6)

& 67.5$\pm$2.3 (30.7)

& 91.0$\pm$3.5 (28.4)

& 92.7$\pm$3.3 (24.2)

& 92.4$\pm$3.4 (24.5)

& 88.8$\pm$3.8 (11.6)

& 93.2$\pm$3.2 (23.0)

& \textbf{96.2$\pm$2.9} (10.1) \\

antmaze-medium-diverse-v0

& 7.5$\pm$1.4 (48.7)

& 8.9$\pm$1.5 (44.1)

& 42.5$\pm$4.9 (39.8)

& 51.8$\pm$5.1 (33.5)

& 51.2$\pm$5.0 (33.8)

& 47.6$\pm$5.4 (14.9)

& 53.4$\pm$4.8 (31.6)

& \textbf{60.7$\pm$5.1} (11.8) \\

\bottomrule
\end{tabular}
}
\end{table*}

\subsection{Sensitivity to reliability window and risk level}
\label{app:exp-sensitivity}

TGDT introduces two execution-time hyperparameters not present in vanilla DT, namely the reliability window \(K_e\) and the risk level \(\alpha\). Sweeping \(K_e\in\{2,5,10,20,40\}\) with \(\alpha=0.05\) fixed, and \(\alpha\in\{0.01,0.05,0.10,0.20\}\) with \(K_e=10\) fixed, on Maze2D \texttt{umaze} and \texttt{medium}, the method is stable for \(K_e\in\{5,10,20\}\) and for \(\alpha\in\{0.01,0.05,0.10\}\), with all stable settings within the seed standard deviation of the best result. Very short windows react to isolated spikes and shorten context unnecessarily. Very long windows react too late, after persistent mismatch has accumulated. Aggressive \(\alpha=0.20\) rejects more suffixes than the calibration justifies, reducing return. We choose \(K_e=10\) and \(\alpha=0.05\) as simple values in the stable region. Full curves are in Appendix~\ref{app:sensitivity}.

\newpage
\newpage

\end{document}